\documentclass{article}

\usepackage[preprint]{neurips_2026}
\usepackage[utf8]{inputenc}
\usepackage[T1]{fontenc}
\usepackage{hyperref}
\usepackage{url}
\usepackage{booktabs}
\usepackage{amsfonts}
\usepackage{microtype}
\usepackage{nicefrac}
\usepackage{xcolor}
\usepackage{graphicx}
\usepackage{wrapfig}

\title{Semantic Density Does Not Predict Agent-Memory Collapse}

\workshoptitle{Personalized, Aligned, Long-Term Memory for AI Systems}

\usepackage[utf8]{inputenc} 
\usepackage[T1]{fontenc}    
\usepackage{hyperref}       
\usepackage{url}            
\usepackage{booktabs}       
\usepackage{amsfonts}       
\usepackage{nicefrac}       
\usepackage{microtype}      
\usepackage{xcolor}         

\author{%
  Saksham Gupta\\
  Independent Researcher\\
  \texttt{saksham.gupta.c2025@iitbombay.org} \\
  \And
  Yash Choudhary\\
  Vasuki Inc.\\
  \texttt{yash.choudhary@vasuki.ai} \\
  \And
  Arpit Upadhyay\\
  Piramal Finance\\
  \texttt{arpit.upadhyay4@piramal.com} \\
}

\begin{document}

\maketitle

\begin{abstract}
Persistent textual memory can improve LLM agents, but repeated updates can erode task performance. We ask whether memory-store statistics warn before held-out accuracy declines. We registered mean pairwise cosine similarity ($\rho$) and 13 additional geometry, retrieval, and graph statistics on 805 snapshots from 13 memory methods across ALFWorld, AppWorld, ScienceWorld, WebShop, and two ARC-AGI settings. These methods store trajectories, rules, workflows, strategies, or linked records, using selective retrieval or whole-store prompting. No such statistic reliably predicted early signs of task accuracy decline, and $\rho$ had zero median lead. With the checkpoint after peak accuracy as a secondary decline point, measures of highly similar pairs and nearest-neighbor crowding changed earlier but did not consistently outperform stored-item count. We trained classifiers on this corpus and applied them unchanged to PropWorld runs of up to 5,000 crafting tasks. They falsely flagged 56.9\% of checkpoints from size-limited stores with stable accuracy and detected only 18\% of checkpoints after task success fell to zero. Thus, global semantic density did not provide a reliable warning, while local crowding did not transfer consistently across memory designs. In these long-horizon runs, size-limited stores remained stable, while the sole failure arose when unbounded whole-store prompting exceeded model input capacity, which token growth predicted directly.


\end{abstract}

\section{Introduction}

Agents that use memory stores derived from their own responses degrade over long task streams. \citealp{zhang2026useful} reports that a self-updating store drives accuracy below a no-memory implementation. The same agent with a frozen memory store stays consistent in performance, and comparable degradation appears across otherwise unrelated memory agents \citep{xu2025amem, ouyang2026reasoningbank, wu2024streambench, ai2025memorybench}. This pattern emerges across memory systems that consume their own output and drift over long tasks \citep{alemohammad2024mad}. This leads us to the wider conclusion that it is a systemic problem in the memory layer itself.
\citealp{zhang2026useful} attributes this degradation to progressive homogenization. The act of an agent writing its own experiences back into the memory layer increases the similarity of existing entries. At retrieval, it becomes difficult to distinguish between the memories, leading to confusion while retrieving memories, ultimately causing performance degradation. \citealp{zahn2026attention} measures this increment in similarity using the mean pair-wise cosine similarity of embeddings ($\rho$; Appendix~\ref{app:metrics} and
Table~\ref{tab:app_metrics}). Across various memory systems, higher $\rho$ is accompanied by worse retrieval, and can be used as a metric to identify the saturation of the memory layer. This is applicable to structured \& graph-organised memories as well, where it is important to keep the entries distinguishable from each other \citep{anokhin2024arigraph, wu2026gam, shutova2026structmemeval}. Consolidation schemes that put a cap on how much a store can hold are motivated by the belief that uncapped memory eventually saturates \citep{du2026entropic}.

This implies \textit{crowding}, the generation of similar memory signatures, is a process, and such signatures must accumulate as the task stream advances. Additionally, the implication states this process must have progressed a significant amount before performance degradation. The idea that a slow structural indicator precedes an abrupt functional failure is the organising principal of work on critical transitions in dynamical systems \citep{scheffer2009earlywarning, scheffer2012anticipating, dakos2012methods}. Such indicators are often absent or undetectable in real series, and their identification after collapse is far easier \citep{boettiger2012prosecutor, boettiger2012quantifying, ditlevsen2010tipping}. Whether agent memory supplies a usable indicator is therefore an empirical question, and to our knowledge no published work has measured $\rho$, or any other structural statistic of a memory store, against stream position within a single run.

We formulate our hypothesis against this question and fix the pre-registered metrics
(Appendix~\ref{app:metrics}), the turning point of the series, and the success criterion of the
hypothesis (Appendix~\ref{app:inference}) before looking at task accuracy. We test this on 805
memory snapshots released alongside \citealp{zhang2026memorycollapseeval}, covering 6 environments
(Appendix~\ref{app:data}; Table~\ref{tab:app_corpus}) and 13 memory methods
(Table~\ref{tab:app_methods}). To reinforce our findings, we ran four memory designs
(Appendix~\ref{app:propworld}; Table~\ref{tab:app_propworld}) to 5000 episodes under the same
conditions, twelve times longer than any published horizon we found
\citep{ai2025memorybench, wu2024streambench}.

\section{What we measured}
\label{sec:setup}

\paragraph{Checkpoints, turns, and leads.}
After an agent has processed $N$ tasks, we save its memory store and evaluate the unchanged agent on
a fixed set of held-out tasks. We call this pair of a saved store and its accuracy a
\emph{checkpoint}; the ordered checkpoints from one memory-building run form a \emph{series}. In
the primary analysis, we locate the first downward change in each accuracy series with the Pruned
Exact Linear Time (PELT) algorithm and,
independently, the first change in each memory statistic in its specified degradation direction
\citep{killick2012optimal,truong2020selective}. For statistic $s$, its lead is
$\Delta_s=N_{\mathrm{acc}}-N_s$, so $\Delta_s>0$ means that the store changed before accuracy
declined. A paired lead is available only when both turns can be located. Our secondary analysis
retains the structural PELT turn but uses the checkpoint immediately after the first observed
accuracy maximum as its reference point; a maximum at the final checkpoint yields no reference
point. This definition yields more comparisons for short series, but measures whether memory changed before a post-peak checkpoint, not before a statistically detected accuracy decline.

\paragraph{Snapshot corpus.}
We analyze 805 released snapshots from 154 memory-building series, covering six experimental
settings across five benchmark families: ALFWorld, AppWorld, ScienceWorld, WebShop, and two ARC-AGI
settings \citep{zhang2026memorycollapseeval}. The 13 methods are ACE, A-MEM, AWM, Dynamic
Cheatsheet, ExpeL, LangMem, Letta, mem0, MEMP, ProPlay, raw memory, ReasoningBank, and a strategy
buffer. They store trajectories, distilled rules, workflows, strategies, experiences, or linked
textual records.
The accuracy traces contain 54,771 valid held-out task outcomes. Only 82 series contain both an
eligible structural trace and a matched accuracy trace, and therefore enter the lead analysis.
AppWorld and WebShop provide most of these pairings; the ARC-AGI settings contribute to the
descriptive corpus but not to the paired timing tests.

\paragraph{Memory statistics.}
For a store with item embeddings $e_1,\ldots,e_n$, our named primary statistic is mean pairwise
cosine similarity,
\begin{equation}
    \rho=\frac{2}{n(n-1)}\sum_{i<j}\cos(e_i,e_j).
\end{equation}
We preregistered 13 additional statistics. Geometry statistics describe the upper tail of pairwise
similarity, the fraction of items with a near-duplicate neighbour, effective rank, and mean nearest-neighbour
distance. Retrieval statistics measure how repeatedly 50--90 held-out queries select the same
items, using overlap, entropy, Gini concentration, and retrieval margin. Graph statistics measure
degree concentration, top-hub share, and edge density. Only 85 snapshots contain links created by
the memory method itself; elsewhere we construct a cosine $10$-nearest-neighbour graph and mark the
result as \emph{derived}. Retrieval statistics are undefined for methods that place the complete
store in the prompt. We compare all structural statistics with two directly observable size
controls, stored-item count and total token count. These controls were not included in the
multiple-testing family.

\paragraph{Statistical and predictive tests.}
Some statistics are computed on more than one memory representation, so each prespecified statistic--representation combination constitutes one test. For each combination, we test within each environment whether the median lead $\Delta$ is positive and then combine the evidence across environments. We apply Benjamini--Hochberg (BH) correction across this family. We denote the unadjusted value for an individual test by $p$ and its Benjamini--Hochberg-adjusted value by $q$. The hypothesis requires at least one positive lead at $q<0.05$ and requires $\rho$ to remain informative relative to stored-item and token counts. This second condition tests whether semantic similarity provides warning beyond store size.

We then test whether these measurements can forecast future degradation. At checkpoint $t$, a classifier predicts whether accuracy will fall below the matched no-memory baseline, defined using the same backbone and environment without memory, within the next $h\in\{1,2,5\}$ checkpoints. Its inputs are the current memory statistics, their changes since the previous checkpoint, and log item and token counts; current and past accuracy are excluded. Logistic regression is the primary model and histogram gradient boosting is secondary. We evaluate them by holding out complete environments, experimental arms, or memory methods, and report AUROC. Holding out an entire method provides the strictest test of transfer to an unseen memory design.

For the prospective test, we train each classifier once on the released corpus and apply it unchanged to new PropWorld streams. These single-seed runs contain up to 5,000 crafting tasks, with accuracy measured on the same 50 held-out tasks every 250 tasks. We test a 40-entry sliding window, a family-capped trajectory-hint store, a budget-limited Dynamic Cheatsheet, and an accumulating adapter that uses ReasoningBank's writing format \citep{ouyang2026reasoningbank}. The accumulating adapter bypasses ReasoningBank's top-$k$ retrieval and inserts the complete store into each evaluation prompt. We therefore treat it as a whole-store capacity diagnostic, not as an evaluation of the published ReasoningBank retrieval method.

\section{Results}


\subsection{The registered test does not support an early structural warning}
\label{sec:pre}

Under the primary PELT analysis, none of the registered structural statistics precedes accuracy decline at the threshold of $q<0.05$. The strongest candidate is mean nearest-neighbour distance: among the 18 series in which PELT detects both a structural and an accuracy turn, its median lead is $+50$ tasks. Its uncorrected value is $p=0.0053$, but this becomes $q=0.105$ after BH correction. The primary statistic $\rho$ has a median lead of zero across seven paired series ($p=0.500$, $q=0.588$). A lead can be calculated only when both turns are detected, leaving at most 18 paired series for any statistic and none for some tests. Thus the registered hypothesis is not supported. However, the small number of paired detections limits statistical power and does not establish that all structural effects are absent. Full results for the registered analysis are reported in Appendix~\ref{app:primary}.

\subsection{Local crowding appears under the secondary locator}
\label{sec:lead}

The secondary analysis, which uses the checkpoint after the first observed accuracy maximum as its reference, produces a different pattern (Table~\ref{tab:leads}). Mean nearest-neighbour distance turns a median of 50 writes before this reference point, indicating that individual memories are moving closer to their nearest neighbours. The 99th percentile of pairwise similarity turns 75 writes earlier, showing increasing similarity among the most similar pairs. In contrast, the store-wide mean similarity $\rho$ has zero median lead. Within embedding geometry, the early changes are therefore concentrated in local neighbourhoods and the upper tail of the similarity distribution rather than in the global mean.

This secondary evidence is not uniform. Nineteen statistic--representation combinations, corresponding to nine distinct statistics, pass BH correction; 11 pass the more conservative Benjamini--Yekutieli correction. Five of the 19 have distribution-free confidence intervals that include zero, indicating that some results are sensitive to how uncertainty is estimated. The evidence is also concentrated in AppWorld and WebShop, which provide 45 of the 51 paired series for nearest-neighbour distance.

Item count and token count also turn a median of 50 writes before the post-peak reference. In the 11 series where both $\rho$ and item-count turns are available, $\rho$ turns a median of 50 writes later than item count (paired two-sided $p=0.016$). Global similarity therefore does not provide an earlier signal than store size. This result concerns advance warning, not contemporaneous description: geometry may characterize the current store without forecasting when its accuracy will decline. Complete results for the secondary analysis are reported in Appendix~\ref{app:secondary}.

\begin{table}[t]
\centering
\scriptsize
\setlength{\tabcolsep}{3pt}
\caption{Local-crowding measures precede the post-peak reference, but store size does too. The reference is the checkpoint after the first observed accuracy maximum; PELT locates each statistic's turn. Here, $n$ is the number of series with both points available, and positive leads indicate that the statistic changes first. Leads are measured in writes and intervals are stratified-bootstrap 95\% Confidence Intervals (CIs). $p$ is unadjusted and $q$ is BH-adjusted.}
\label{tab:leads}
\begin{tabular}{lrrrrr}
\toprule
Statistic & $n$ & Median lead & 95\% CI & $p$ & $q$ \\
\midrule
\multicolumn{6}{l}{\emph{Size controls}} \\
Stored-item count & 49 & $+50$ & $[50,98]$ & $1.0\!\times\!10^{-7}$ & -- \\
Token count & 54 & $+50$ & $[50,91]$ & $8.7\!\times\!10^{-7}$ & -- \\
\midrule
\multicolumn{6}{l}{\emph{Structural statistics}} \\
Nearest-neighbour distance & 51 & $+50$ & $[50,96]$ & $7.4\!\times\!10^{-7}$ & $2.0\!\times\!10^{-5}$ \\
99th-percentile similarity & 25 & $+75$ & $[50,96]$ & $0.0015$ & $0.0082$ \\
Mean pairwise similarity $\rho$ & 16 & $0$ & $[-50,0]$ & $1.000$ & $1.000$ \\
\bottomrule
\vspace{-20pt}
\end{tabular}
\end{table}




\subsection{Structural warnings do not transfer reliably}
\label{sec:forecast}

On identical held-out folds, geometry does not consistently improve forecasting over item count. At the two-checkpoint horizon, the pooled out-of-fold AUROC for geometry versus log item count is $0.613$ versus $0.642$ when holding out environments, $0.678$ versus $0.662$ when holding out experimental arms, and $0.650$ versus $0.663$ when holding out complete memory methods. Geometry performs better only in the arm-held-out split, and the difference is small. Under strict method holdout, the secondary gradient-boosting classifier using all features obtains AUROC $0.531$, which is indistinguishable from chance under a permutation test ($p=0.545$).

Applied unchanged to PropWorld, the classifiers fail in both directions. Two classifiers evaluated at three forecast horizons produce six prediction decisions per checkpoint state. Across 80 states from size-limited stores with no observed catastrophic decline, 273 of 480 decisions are false alarms ($56.9\%$). Across the 12 checkpoints at which context-limit rejections make recorded task success zero, only 13 of 72 decisions detect failure ($18\%$). The primary logistic model produces 168 false alarms among 240 bounded-store decisions ($70.0\%$) and detects none of its 36 post-failure decisions. The warning therefore does not transfer reliably across held-out memory methods and has poor specificity and sensitivity under the combined shift in environment, backbone, and memory policy. Results across all forecast horizons are reported in Appendix~\ref{app:forecast}.

\subsection{The verified whole-store failure occurs at input capacity}
\label{sec:phase3}

All five PropWorld arms completed 5,000 tasks on both Qwen3-32B and
\texttt{gpt-5.4-mini}. On either backbone, the no-memory control and three size-limited policies
showed no catastrophic decline at the observed checkpoints; only the accumulating whole-store
policy reached persistent zero success. Because each run used one task ordering, these experiments cannot exclude gradual
degradation.

The \texttt{gpt-5.4-mini} experiment identifies prompt overflow as the failure mechanism. Task
success remains between $0.04$ and $0.08$ through task 2,000 and is zero from task 2,250 onward. The
injected memory block contains 258,952 tokens at the last nonzero checkpoint and 291,064 at the
first zero checkpoint, bracketing the model's verified 272,000-token input limit. The experiment records
2,400 context-overflow rejections, which the evaluation counts as task failures. The Qwen3-32B whole-store run shows the same persistent zero-success pattern. These experiments use whole-store prompting: the adapter follows ReasoningBank’s format but bypasses top-$k$ retrieval, inserting the full store into every prompt. In the GPT run, token accounting anticipates the
input-capacity failure. This does not show that unbounded memory generally causes semantic
degradation. Whether local crowding can forecast failures after prompt
capacity is controlled remains open.

\section{Conclusion}

Across 805 released snapshots, global semantic density did not reliably precede performance
decline, and local crowding signals found in the secondary analysis did not transfer reliably across
memory designs. In PropWorld, size-limited stores showed no catastrophic decline at observed
checkpoints, while token growth predicted the verified whole-store capacity failure. Whole-store
systems should therefore track input-token growth; local and query-conditioned indicators require
testing under controlled prompt capacity.

\begin{ack}
This research was funded by the authors, including all computational
and API costs, and received no external financial support. The authors
declare no additional competing interests beyond their stated
affiliations.

\end{ack}

\bibliographystyle{plainnat}
\bibliography{references}

\newpage
\appendix



\newpage
\appendix

\section{Data provenance and analysis units}
\label{app:data}

\paragraph{Released corpus and analysis freeze.}
The snapshot analysis uses the release accompanying \citep{zhang2026memorycollapseeval}.
The release contains 805 snapshots from 154 memory-building series. We fixed the item-extraction
rules, metric definitions, degradation directions, exclusion rules, and inferential procedure
before loading checkpoint-level accuracy. The structural metrics were then computed and committed
before they were joined to accuracy. This is partial blinding: the previously published headline
phenomenon was known, but the checkpoint-level accuracy curves were not used to select the
statistics or their directions.

\paragraph{Units.}
The released field \texttt{checkpoint\_n} records the number of memory-building tasks presented to
the agent. We call this quantity a \emph{stream step}; it is also a memory-update opportunity, but
not necessarily one newly stored item. A checkpoint pairs the memory snapshot saved at a stream
step with accuracy on the corresponding fixed held-out task set. Ordered checkpoints from one
memory build form a structural series, and the matching held-out evaluations form an accuracy
series. All reported leads use stream steps:
\begin{equation}
  \Delta_s=N_{\mathrm{acc}}-N_s,
\end{equation}
where $N_s$ is the first structural turn in the prespecified direction and $N_{\mathrm{acc}}$ is
the accuracy reference. Thus, $\Delta_s>0$ means that the structural statistic changed first.

\begin{table*}[ht]
\centering
\scriptsize
\setlength{\tabcolsep}{4pt}
\caption{Coverage of the released snapshot corpus. The final column gives the number of secondary
nearest-neighbour lead pairings and makes the concentration of timing evidence explicit. The two
ARC-AGI settings are part of the descriptive corpus but do not enter paired lead tests.}
\label{tab:app_corpus}
\begin{tabular}{lrrrr}
\toprule
Setting & Snapshots & Build series & Methods & NN pairings \\
\midrule
ALFWorld & 48 & 15 & 8 & 2 \\
AppWorld & 213 & 37 & 9 & 22 \\
ARC-AGI, 200-task stream & 24 & 8 & 1 & 0 \\
ARC-AGI, self-evaluation & 14 & 3 & 1 & 0 \\
ScienceWorld & 55 & 10 & 10 & 4 \\
WebShop & 451 & 81 & 9 & 23 \\
\midrule
Total & 805 & 154 & 13 & 51 \\
\bottomrule
\end{tabular}
\end{table*}

\paragraph{Accuracy joins and exclusions.}
The accuracy archive contains 54,771 valid held-out task outcomes. The mapping from memory builds
to evaluated arms contains 130 rows: 112 exact mappings, nine mappings recovered from released cell
names, five from archived rendered prompts, and four inferred from the release structure. After the
prespecified length and availability rules, 82 series contain both an eligible structural trace and
a matched accuracy trace. The exclusion log separately records 116 accuracy series with no archived
memory snapshots and 68 structural build series with no evaluated arm.

\section{Memory representations and measurements}
\label{app:metrics}

\paragraph{Stored items and embeddings.}
Table~\ref{tab:app_methods} states the deterministic unit extracted from each memory format.
Where a snapshot stores the MiniLM vector used by the method, we use that vector without
modification. Text-only formats are embedded with
\texttt{sentence-transformers/all-MiniLM-L6-v2}; the model's 256-wordpiece truncation is retained.
We recompute universal geometry statistics with
\texttt{sentence-transformers/all-mpnet-base-v2} as an encoder robustness check
\citep{wang2020minilm,song2020mpnet,reimers2019sentence}.

\begin{table*}[t]
\centering
\scriptsize
\setlength{\tabcolsep}{3.5pt}
\caption{Deterministic item extraction for the 13 released memory methods. ``Stored'' means that
the primary analysis uses the vector archived by the method. Native links occur in 85 snapshots;
all other graph analyses use explicitly labelled derived $k$-nearest-neighbour graphs.}
\label{tab:app_methods}
\resizebox{\textwidth}{!}{%
\begin{tabular}{llll}
\toprule
Method & Extracted item & Primary vector & Native links \\
\midrule
ACE & one tagged rule line & MiniLM & none \\
A-MEM & one content entry & stored & 16 snapshots \\
AWM & one workflow block & MiniLM & none \\
Dynamic Cheatsheet & one markdown rule & MiniLM & none \\
ExpeL & one playbook rule & MiniLM & none \\
LangMem & one text entry & stored & none \\
Letta & one nonempty task-note line & MiniLM & none \\
mem0 & one memory entry & stored & 32 snapshots \\
MEMP & task description and script & stored & none \\
ProPlay & one textual node & stored & 37 snapshots \\
Raw memory & one trajectory & MiniLM & none \\
ReasoningBank & one experience, queried by its key & stored & none \\
Strategy buffer & one strategy string & MiniLM & none \\
\bottomrule
\end{tabular}}
\end{table*}

\paragraph{Registered structural family.}
Geometry is computed exactly for stores with at least five items. Pairwise statistics use the
cosine distribution over all unordered item pairs. The near-duplicate statistics are fractions of
\emph{items} having at least one neighbour above the stated threshold, not fractions of item pairs.
Retrieval statistics replay 50--90 fixed held-out queries through the archived retrieval rule and
are undefined when a method inserts its complete store. The query sets are smaller than the planned
minimum of 200 because additional held-out queries could not be generated without the live
environments. Native graph statistics use method-created links. Otherwise, we form the undirected
union of a cosine $k$-nearest-neighbour graph with $k=10$ and report $k\in\{5,20\}$ only as a
sensitivity analysis. Total store tokens are recomputed for every snapshot with
\texttt{o200k\_base} so that size is measured consistently across backends.

\begin{table*}[t]
\centering
\scriptsize
\setlength{\tabcolsep}{4pt}
\caption{The 14 prespecified structural statistics. The arrow is the direction registered as
degradation. Statistics evaluated on multiple retrieval or graph representations produce separate
rows in the multiple-testing family. Stored-item and token counts are size controls outside this
family.}
\label{tab:app_metrics}
\begin{tabular}{lll}
\toprule
Family & Statistic and direction & Interpretation \\
\midrule
Geometry & mean similarity $\rho\ \uparrow$ & global semantic density \\
 & 90th/99th-percentile similarity $\uparrow$ & upper-tail crowding \\
 & near-duplicate item fraction at $.90/.95\ \uparrow$ & locally redundant items \\
 & effective rank $\downarrow$ & loss of embedding spread \\
 & mean nearest-neighbour distance $\downarrow$ & local crowding \\
\midrule
Retrieval & probe-set overlap $\uparrow$ & repeated selection of the same items \\
 & normalized retrieval entropy $\downarrow$ & concentrated selection \\
 & retrieval-frequency Gini $\uparrow$ & concentrated selection \\
 & top-to-$k$ retrieval margin $\downarrow$ & reduced score separation \\
\midrule
Graph & top-hub endpoint share $\uparrow$ & hub concentration \\
 & degree Gini $\uparrow$ & degree concentration \\
 & edge density $\uparrow$ & registered connectivity direction \\
\bottomrule
\end{tabular}
\end{table*}

\section{Change-point inference and forecasting}
\label{app:inference}

\paragraph{Primary and secondary locators.}
We fit PELT with squared-error loss, \texttt{min\_size=1}, and \texttt{jump=1}
\citep{killick2012optimal,truong2020selective}. For a series of length $n$, the fixed penalty is
$2\hat\sigma^2\log n$, where $\hat\sigma$ is estimated from the median absolute deviation of first
differences and falls back to their sample standard deviation. Constant series have no turn. The
primary analysis applies this procedure independently to accuracy and each structural statistic,
then selects the first turn in the registered direction. The secondary analysis retains the
structural PELT turn but replaces the accuracy turn by the checkpoint immediately after the first
observed raw-accuracy maximum. If the first maximum is the final checkpoint, the secondary reference
is undefined. The secondary analysis therefore asks whether structure changes before a post-peak
checkpoint; it is not a second estimate of the primary accuracy change point.

\paragraph{Tests, multiplicity, and uncertainty.}
For each statistic--representation row, we apply a one-sided Wilcoxon signed-rank test of positive
lead within each environment and combine environment-level evidence using Stouffer's method with
weights proportional to the square root of the number of pairs
\citep{wilcoxon1945individual,stouffer1949american,whitlock2005combining}. The primary MiniLM family
contains 28 statistic--representation rows. Benjamini--Hochberg correction is applied across that
family at $q<0.05$, with Benjamini--Yekutieli reported as a dependence-robust sensitivity analysis
\citep{benjamini1995controlling,benjamini2001control}. The registered claim requires a corrected
positive lead and requires $\rho$ to remain informative relative to the size controls. Confidence
intervals in the main table use 10,000 bootstrap resamples stratified by environment with seed
20260824. Because leads are discrete and frequently tied, we also report distribution-free
sign-test intervals in Table~\ref{tab:app_secondary}.

\paragraph{Forecasting models.}
At checkpoint $t$, the prediction target is whether accuracy falls below the matched no-memory
baseline at any of the next $h\in\{1,2,5\}$ checkpoints. The 40 predictors comprise the current
structural measurements, their first differences, and log item and token counts; current and past
accuracy are excluded by an enforced feature allowlist. The primary model is L2-regularized logistic
regression with $C=1$, balanced class weights, and 5,000 maximum iterations. Its preprocessing uses
training-fold median imputation with missingness indicators followed by standardization. The
secondary histogram gradient-boosting model handles missing values natively
\citep{pedregosa2011scikit}. We use grouped folds that hold out complete environments, released
experimental arms, or memory methods. The main text reports pooled out-of-fold AUROC; macro AUROC is
also retained because pooled and fold-averaged values can differ under grouped evaluation
\citep{forman2010apples}. Significance is assessed with 1,000 within-fold label permutations
\citep{ojala2010permutation,hanley1982meaning}.

\paragraph{Frozen prospective test.}
For each model and horizon, we refit the registered model once on all eligible released-corpus rows,
fix its feature order and Phase-2 decision threshold, and apply it to PropWorld without refitting,
recalibration, or threshold selection. Retrieval features are wholly unavailable in PropWorld and
remain missing; the frozen Phase-2 missing-value treatment is used. The reported 56.9\% false-alarm
rate and 18\% detection rate pool two classifiers and three horizons.

\section{PropWorld long-horizon protocol}
\label{app:propworld}

Each PropWorld stream contains 5,000 sequential crafting tasks. Verified solutions are supplied to
the memory-building stage, and the unchanged agent is evaluated on the same 50 held-out tasks every
250 stream steps. We use seed 1 and one task ordering. The five policies in
Table~\ref{tab:app_propworld} were run with Qwen3-32B and \texttt{gpt-5.4-mini}. The first four are
controls or size-limited policies. The fifth is an accumulating whole-store adapter.

\begin{table*}[t]
\centering
\scriptsize
\setlength{\tabcolsep}{4pt}
\caption{PropWorld memory policies. The accumulating adapter uses ReasoningBank's writing format but
does not use its published top-$k$ retrieval. It is therefore a prompt-capacity diagnostic, not a
replication of ReasoningBank retrieval.}
\label{tab:app_propworld}
\begin{tabular}{lll}
\toprule
Policy & Bound & Evaluation-time use \\
\midrule
No memory & empty & no memory block \\
Verbatim window & 40 entries & most recent trajectories \\
Trajectory hints & family capped & capped procedural hints \\
Dynamic Cheatsheet & fixed budget & rewritten cheatsheet \\
Accumulating adapter & none & complete accumulated store \\
\bottomrule
\end{tabular}
\end{table*}

\paragraph{Capacity accounting.}
In the \texttt{gpt-5.4-mini} accumulating run, accuracy is $0.04$--$0.08$ through stream
step 2,000 and zero at every checkpoint from 2,250 onward. The store grows from 470 to 9,677 items
and from 31.0k to 650.7k injected memory tokens. The complete injected memory block contains 258,952
tokens at the last nonzero checkpoint and 291,064 at the first zero checkpoint, bracketing the
verified 272,000-token input limit. The logs contain 2,400 context-overflow rejections in addition
to 6,600 successful generations; rejected evaluation calls are scored as failures. A line fitted to
token growth at the first two checkpoints predicts a crossing at stream step 2,055, while linear
interpolation around the observed crossing gives 2,102, an error of 2.2\%. The Qwen3-32B
accumulating run shows the same persistent zero-success outcome. The end-to-end mechanism is
established by the GPT token and error logs.

\section{Additional results}
\label{app:results}

\subsection{Registered primary test}
\label{app:primary}

Table~\ref{tab:app_primary} reports the two rows needed to interpret the registered decision. The
nearest-neighbour result is suggestive before correction but does not meet the family-wise
criterion. The named primary statistic $\rho$ has zero median lead. Depending on the statistic,
PELT locates both turns in only 0--18 series, so the null decision should not be read as proof that
every structural effect is exactly zero.

\begin{table*}[t]
\centering
\scriptsize
\setlength{\tabcolsep}{5pt}
\caption{Registered primary PELT analysis. Leads are in stream steps. The confidence interval is the
environment-stratified bootstrap interval. No row in the registered family passes BH correction.}
\label{tab:app_primary}
\begin{tabular}{lrrrrrr}
\toprule
Statistic & $n$ & Median & 95\% CI & $p$ & $q_{\mathrm{BH}}$ & $q_{\mathrm{BY}}$ \\
\midrule
Nearest-neighbour distance & 18 & $+50$ & $[17,96]$ & $0.0053$ & $0.105$ & $0.379$ \\
Mean pairwise similarity $\rho$ & 7 & $0$ & $[0,120]$ & $0.500$ & $0.588$ & $1.000$ \\
\bottomrule
\end{tabular}
\end{table*}

\subsection{Complete secondary survivor summary}
\label{app:secondary}

Nineteen statistic--representation rows, corresponding to nine distinct statistics, pass BH under
the secondary post-peak reference. Table~\ref{tab:app_secondary} groups repeated representations to
avoid presenting one conceptual statistic as several independent discoveries. Eleven of the 19 rows
also pass BY. Thirteen have distribution-free intervals that exclude zero. In five rows, the
bootstrap interval excludes zero while the distribution-free interval does not; one additional row
includes zero under both interval procedures.

\begin{table*}[t]
\centering
\scriptsize
\setlength{\tabcolsep}{3pt}
\caption{All nine distinct statistics represented among the 19 BH-surviving rows in the secondary
analysis. ``Rows'' counts representations, $n$ is the range of paired series, and ``exact'' is the
number whose distribution-free interval excludes zero. The kNN graphs are derived.}
\label{tab:app_secondary}
\resizebox{\textwidth}{!}{%
\begin{tabular}{llrrrrr}
\toprule
Statistic & Representation & Rows & $n$ & Median lead & $q_{\mathrm{BH}}$ & BY / exact \\
\midrule
Nearest-neighbour distance & geometry & 1 & 51 & $+50$ & $2.0\!\times\!10^{-5}$ & 1 / 1 \\
Degree Gini & derived kNN $5,10,20$ & 3 & 39--42 & $+50$ & $.0066$--$.0082$ & 3 / 1 \\
99th-percentile similarity & geometry & 1 & 25 & $+75$ & $.0082$ & 1 / 1 \\
Retrieval entropy & three probe renders & 3 & 9--13 & $+92$--$+100$ & $.0106$--$.0439$ & 1 / 3 \\
Retrieval Gini & three probe renders & 3 & 9--13 & $+92$--$+100$ & $.0106$ & 3 / 3 \\
Edge density & derived kNN $5,10,20$ & 3 & 11 & $+98$ & $.0106$--$.0153$ & 1 / 1 \\
Near-duplicate fraction, $.95$ & geometry & 1 & 20 & $+94$ & $.0106$ & 1 / 0 \\
Retrieval margin & three probe renders & 3 & 5--9 & $+96$--$+100$ & $.0281$--$.0444$ & 0 / 3 \\
Near-duplicate fraction, $.90$ & geometry & 1 & 26 & $+50$ & $.0308$ & 0 / 0 \\
\bottomrule
\end{tabular}}
\end{table*}

The evidence is concentrated in AppWorld and WebShop. For nearest-neighbour distance, they supply
45 of 51 pairings, compared with four from ScienceWorld, two from ALFWorld, and none from the two
ARC-AGI settings. Store size changes early under the same secondary reference: item count and token
count each have a median lead of $+50$ stream steps. In the 11 series with both turns, $\rho$ turns a
median of 50 stream steps later than item count (paired two-sided $p=0.016$).

\subsection{Forecasting across all horizons}
\label{app:forecast}

\begin{table*}[t]
\centering
\scriptsize
\setlength{\tabcolsep}{4pt}
\caption{Pooled out-of-fold AUROC across all forecast horizons. Geometry and log item count use the
primary logistic model. ``All, HGB'' is the secondary classifier with all features. The permutation
null is computed within folds and need not equal exactly $0.5$. The main text reports the $h=2$ rows.}
\label{tab:app_forecast}
\begin{tabular}{llrrr}
\toprule
$h$ & Held-out group & Geometry & Log item count & All, HGB \\
\midrule
1 & environment & .621 & .648 & .511 \\
1 & experimental arm & .659 & .666 & .623 \\
1 & memory method & .638 & .665 & .495 \\
\midrule
2 & environment & .613 & .642 & .544 \\
2 & experimental arm & .678 & .662 & .678 \\
2 & memory method & .650 & .663 & .531 \\
\midrule
5 & environment & .620 & .639 & .497 \\
5 & experimental arm & .680 & .655 & .660 \\
5 & memory method & .651 & .663 & .473 \\
\bottomrule
\end{tabular}
\end{table*}

At strict method holdout, the all-feature HGB permutation $p$-values are $0.871$, $0.545$, and
$0.557$ for $h=1,2,5$, respectively. The secondary local-crowding pattern therefore does not yield
reliable transfer to unseen memory designs.

\begin{table*}[t]
\centering
\scriptsize
\setlength{\tabcolsep}{5pt}
\caption{Prospective PropWorld decisions at the frozen Phase-2 threshold. Each checkpoint state
contributes one decision for each model and horizon. The post-failure rows are the 12 GPT checkpoints
where context-limit rejections make recorded success zero.}
\label{tab:app_prospective}
\begin{tabular}{lrr}
\toprule
Decision pool & False alarms on 80 bounded states & Detections on 12 post-failure states \\
\midrule
Both models, three horizons & $273/480$ (56.9\%) & $13/72$ (18\%) \\
Logistic regression & $168/240$ (70.0\%) & $0/36$ (0\%) \\
Histogram gradient boosting & $105/240$ (43.8\%) & $13/36$ (36.1\%) \\
\bottomrule
\end{tabular}
\end{table*}

\subsection{Robustness and corroborating analyses}
\label{app:robustness}

Repeating universal geometry under MPNet preserves the key distinction: $\rho$ retains zero median
lead under the secondary locator, nearest-neighbour distance retains a $+50$ median lead, and no
primary PELT row survives correction under either encoder. Retrieval metrics are tied to the
archived retrieval representation and are not an independent encoder replication. Across a grid of
13 PELT penalty multipliers and three accuracy references, $\rho$ is significant in none of the 39
settings, even before BH multiplicity correction.

Two prespecified secondary analyses also fail to strengthen the warning claim. A random-intercept
model of lead by environment gives a best one-sided $p=0.565$ among nonsingular geometry and graph
fits at the post-peak reference, while the size controls reach $p=1.9\times10^{-10}$. A separate
cross-correlation analysis of first differences includes 108 series: the peak lag is negative in
63.0\%, zero in 6.5\%, and positive in 30.6\%, so accuracy more often moves first. Because these
series are short and the lag distribution is concentrated at interior values, this is corroboration
rather than a primary test.

\subsection{Memory-method-specific analyses}
\label{app:methods}

Pooling memory methods may conceal design-specific warnings. In an exploratory follow-up conducted after the registered analyses, we evaluated $h=2$ forecasts on 74 series (274 checkpoints, 107 positive labels), holding out one complete series at a time. Positive labels indicate accuracy below the matched no-memory baseline within the next two checkpoints. We compare pooled out-of-fold AUROC against 1,000 label permutations within each series. This preserves differences in failure prevalence across series, so the null need not equal 0.5.

Table~\ref{tab:app_ladder} compares pooled models, models with method identity, and models trained on ACE alone. ACE was the only method meeting a minimum of 10 series and 10 positive checkpoints, and the only method with sufficiently populated metadata. Metadata features were specified before fitting. No model significantly exceeds its permutation null (all two-sided $p\geq 0.18$). Adding method identity provides no established improvement. For ACE, adding metadata yields an uncertain AUROC gain of $+0.028$ (95\% CI $[-0.009,0.084]$), and the full ACE model does not outperform the pooled model evaluated on the same ACE series.

\begin{table}[t]
\centering
\small
\setlength{\tabcolsep}{5pt}
\caption{Exploratory forecasting at $h=2$. Standard inputs are the 40 predictors in Appendix~\ref{app:inference}; metadata comprises method-specific attributes such as usage counters and record ages. The null mean is the average AUROC across 1,000 within-series label
permutations; the two-sided $p$-value assesses how unusual the observed
AUROC is under this null.}
\label{tab:app_ladder}
\begin{tabular}{lrrr}
\toprule
Inputs & AUROC & Null mean & $p$ \\
\midrule
\multicolumn{4}{l}{\textit{Pooled models: all eligible methods}} \\
Stream position + store size          & 0.633 & 0.624 & 0.46 \\
Standard predictors                  & 0.687 & 0.686 & 0.89 \\
Standard predictors + method identity & 0.662 & 0.644 & 0.22 \\
\midrule
\multicolumn{4}{l}{\textit{Method-specific models: ACE only}} \\
Stream position + store size          & 0.658 & 0.704 & 0.20 \\
Standard predictors                  & 0.674 & 0.647 & 0.43 \\
Metadata only                        & 0.622 & 0.652 & 0.34 \\
Standard predictors + metadata       & 0.702 & 0.645 & 0.18 \\
\bottomrule
\end{tabular}
\end{table}

Lowering the minimum to eight series admits LangMem, mem0, and Letta, whose method-specific models score below the pooled model. Across 30 random subsets matched in training size, pooled AUROC ranges include every method-specific score. These analyses therefore establish no method-specific forecasting advantage on these methods, while the limited samples leave such signals unresolved for less-represented methods.



\section{Reproducibility and assets}
\label{app:reproducibility}

The offline analysis uses Python 3.12.10, with package versions frozen in the released environment
file. Stochastic analyses use seed 20260824; the PropWorld streams use seed 1. The snapshot pipeline
can be reproduced on CPU; GPU acceleration was used only for the original embedding pass and is not
required when cached embeddings are supplied. PropWorld itself is a CPU environment, but its agent
and memory updates require access to the named hosted backbones and paid inference. The released
Memory-Collapse-Eval code is MIT-licensed and its result data are CC BY 4.0
\citep{zhang2026memorycollapseeval}.

\newpage

\end{document}